\documentclass[11pt]{article}

\usepackage[a4paper, margin=2cm]{geometry}
\usepackage[width=.88\textwidth]{caption}
\usepackage{graphicx}
\usepackage{xcolor}
\usepackage{booktabs}
\usepackage{hyperref}
\usepackage[vietnamese, english]{babel}

\usepackage[font=small,labelfont=bf]{caption}
\usepackage{scrextend}
\deffootnote[.65in]{0pt}{1.6em}{\makebox[.3in][l]{\hspace{10pt}\thefootnotemark}}

\begin{document}

\title{Fake Advertisements Detection \\ Using Automated Multimodal Learning: \\ A Case Study for Vietnamese Real Estate Data}

\author{Duy Nguyen$^{1}$ \qquad Trung T. Nguyen$^{2}$ \qquad Cuong V. Nguyen$^{3}$ \\[15pt]
$^{1}$Datal, Ho Chi Minh City, Vietnam \\
\texttt{duyndv@datal.com.vn} \\[10pt]
$^{2}$Department of Data Science \\ 
University of Management and Technology \\
Ho Chi Minh City, Vietnam \\
\texttt{trung.nguyentan@umt.edu.vn} \\[10pt]
$^{3}$Department of Mathematical Sciences \\
Durham University, United Kingdom \\
\texttt{viet.c.nguyen@durham.ac.uk}
}
\date{}

\maketitle
\date{}

\abstract{The popularity of e-commerce has given rise to fake advertisements that can expose users to financial and data risks while damaging the reputation of these e-commerce platforms. For these reasons, detecting and removing such fake advertisements are important for the success of e-commerce websites. In this paper, we propose FADAML, a novel end-to-end machine learning system to detect and filter out fake online advertisements. Our system combines techniques in multimodal machine learning and automated machine learning to achieve a high detection rate. As a case study, we apply FADAML to detect fake advertisements on popular Vietnamese real estate websites. Our experiments show that we can achieve 91.5\% detection accuracy, which significantly outperforms three different state-of-the-art fake news detection systems.}

\section{Introduction}
\label{sec:intro}

With the popularity of e-commerce, online advertisements and marketplaces have been used widely by both sellers and buyers. These e-commerce marketplaces are growing rapidly, supporting the sale and purchase of different products such as real estates, household goods, labor-intensive equipments, etc. With the increased access to these marketplaces, there is also an increase in the number of fake advertisements on these platforms that target vulnerable buyers to obtain their personal information or to trick them into buying fake products~\cite{sadeghpour2021ads}. These fake advertisements severely affect both the genuine sellers and buyers, and potentially undermine the creditability of the e-commerce websites if left undetected.

In this work, we will develop a novel end-to-end machine learning system, called \emph{Fake Advertisements Detection using Automated Multimodal Learning} (FADAML), to detect and filter out fake online advertisements. Our system combines techniques in multimodal machine learning~\cite{baltruvsaitis2018multimodal, nguyen2021multimodal} and automated machine learning (AutoML)~\cite{hutter2019automated, zoller2021benchmark} to improve the system's detection rate. To our knowledge, this is the first work that combines multimodal and automated machine learning for fake advertisement detection. By utilizing multimodal learning, we can leverage both free-form text-based inputs as well as specialized handcrafted features to improve the performance of our system. Additionally, by incorporating AutoML, we can train several powerful machine learning models and combine them to further enhance the performance.

In general, our FADAML system consists of three main components: a data crawler and preprocessor, a multimodal feature extractor, and an AutoML system. The first two components extract and transform raw advertisement texts into refined multimodal features through several processing steps. These refined features are then fed into the AutoML component to train and select the best model for the data, which would be used to filter new advertisements. One distinctive feature of our approach when compared with previous work on multimodal fake news detection~\cite{wang2018eann, palani2022cb, peng2022effective, zhao2023fake} is that we use AutoML to train our model. This simplifies the implementation of our method and makes it highly flexible and applicable. Furthermore, in our work, instead of using generic multimodal information, we work with domain experts to select highly usable multimodal features for our model.

As a case study, we apply our system to detect fake advertisements on Vietnamese real estate websites. The residential real estate market in Vietnam is valued about \$7.3B in 2022 with the majority of real estate agents relying on e-commerce websites to increase transactions~\cite{vietnam_stats}. Furthermore, Vietnamese is a low-resource language that is usually very challenging to work with due to (1) the relatively smaller corpus, (2) the difference in word structures compared to English, and (3) the difference between the corpus used to pre-train language models (i.e., Vietnamese Wikipedia's articles) and that of the downstream tasks (e.g., real estate advertisements)~\cite{nguyen2020phobert}. In this paper, we show that our system can work well for the Vietnamese language.

We implement and test our FADAML system thoroughly using a real-world data set collected from five popular Vietnamese real estate websites. Our experiments show that FADAML can achieve 91.5\% accuracy when detecting fake advertisements, significantly outperforming three state-of-the-art fake news detection baseline methods. To gain a better understanding into our system, we inspect the individual models trained by AutoML and also investigate the feature importance scores of our final model. Finally, we conduct an ablation study to explore the effects of different multimodal feature sets on the performance of our system.

In summary, our paper makes the following contributions. (1) We develop FADAML, a novel end-to-end machine learning system to detect and filter out fake online advertisements. Our system combines techniques in multimodal machine learning and automated machine learning to perform this task. (2) We show how our system can be applied to detect fake advertisements on popular Vietnamese real estate websites. We also empirically show that the system can achieve good detection accuracy for this problem.

\section{Related Work}
\label{sec:related_work}

\subsection{Fake Advertisement Detection}

Online fake advertisement detection~\cite{tran2011spam, mccormick2013discovering, al2020ads, alsaleh2018heuristic} is an important research topic whose solutions can potentially help protect e-commerce websites' users from potential scams. Previous methods to detect such fake advertisements range from using simple classifiers~\cite{tran2011spam} and data mining~\cite{mccormick2013discovering} to heuristic statistics~\cite{alsaleh2018heuristic} and combinations of available online services~\cite{al2020ads}. These previous works, however, focused on the English language and did not use the state-of-the-art advancements in machine learning, such as deep learning, multimodal machine learning, or automated machine learning. In this work, we approach this problem with a solution that combines multimodal and automated machine learning, the two useful frameworks for this type of data. Additionally, our work focuses on advertisements in the Vietnamese language, which received no prior attention. 

The fake advertisement detection problem here is also related to fake news detection~\cite{shu2017fake} and misinformation detection~\cite{islam2020deep} on online social networks. An approach for this problem was proposed in~\cite{antoun2020state} that uses contextualized word embeddings from a pre-trained ELMo model~\cite{peters2018deep} to input into a Bi-LSTM with an attention layer and a softmax classifier for final predictions. Another solution was proposed in~\cite{kaliyar2020fndnet} that introduced FNDNet, which leverages parallel convolutional layers with varying kernel sizes and word embeddings to extract semantic relationships. Both of these methods can achieve remarkable performance on fake news detection; thus, we use them as baselines in this paper to compare with our system.

Fake news detection in non-English languages was also considered in some previous work. In~\cite{faustini2020fake}, three language groups (i.e., Germanic, Latin, and Slavic) were considered and a text-feature method was suggested for identifying fake news. Another work~\cite{posadas2019detection} created a corpus of both fake and real news in Spanish and detected fake news by using linguistically motivated features. For German, \cite{mattern2021fang} developed FANG-COVID, a large-scale benchmark dataset for fake news detection related to the COVID-19 pandemic. Using an explainable textual and social context-based model, they proposed a method that can achieve comparable results to the black box model solely relying on BERT-embeddings. For the Portuguese language, \cite{monteiro2018contributions} investigated the linguistic characteristics of fake news and applied machine learning techniques to achieve significant results. However, unlike our approach, none of these work employs AutoML in their solutions.

\subsection{Vietnamese Natural Language Processing}

Our work is related to natural language processing for the Vietnamese language. In recent years, there have been several works investigating different tasks for this language, such as part-of-speech tagging \cite{phonlp, tran2020improving}, punctuation prediction~\cite{pham2019punctuation, tran2021efficient}, named entity recognition~\cite{pham2015semi, covid19, phan2022ner2ques, phan2022named}, and dependency parsing~\cite{phonlp, do2022adapting}. For Vietnamese, a robust pre-trained transformer-based language model, called PhoBERT~\cite{nguyen2020phobert}, has also been developed and applied to the part-of-speech tagging, named entity recognition, and dependency parsing problems above. In this paper, we will use PhoBERT to obtain the embeddings for the named entity recognition module of our system.

\subsection{Real Estate Data Analysis}

Real estate data analysis has gained much attention recently with the availability of data and analysis tools~\cite{grybauskas2021predictive, fuerst2020real, gale2023optimization}. Among these analyses, property price estimation is an important problem and several approaches have been proposed for this problem that include vision-based approach~\cite{vision}, eager learning~\cite{valuation}, graph convolutional neural network~\cite{luce}, etc. In terms of Vietnamese real estate data, there have been previous works that develop an end-to-end information extraction platform~\cite{end-to-end-ner} and deep learning models for named entity recognition~\cite{ner}. However, the previous work has not considered the fake advertisement detection problem that we consider in this paper. Furthermore, to the best of our knowledge, our paper is also the first to adopt automated and multimodal machine learning to analyze real estate data.

\subsection{Automated Machine Learning}

Our paper utilizes AutoML as a major component in our proposed system. AutoML~\cite{hutter2019automated} is a field that aims to automatically determine the best machine learning approach given a dataset for an application, and then train a model with the best hyperparameter setting using this approach. Researchers and practitioners of AutoML have developed several effective and useful frameworks, including Auto-WEKA~\cite{autoweka}, Auto-Sklearn~\cite{autosklearn}, TPOT~\cite{tpot}, H2O~\cite{h2o}, AutoGluon~\cite{agtabular}, etc. These frameworks are enhancing the practical applicability of AutoML on data-driven model building and automated decision making. Among them, AutoGluon can often achieve a remarkable accuracy on raw data while making a better use of the allotted training time through assembling several models and stacking them in layers~\cite{benchmarking}. For this reason, we use AutoGluon to implement the AutoML component in our system.

\subsection{Multimodal Machine Learning}

One of the main strengths of our proposed system is to combine texts in the advertisements with hand-crafted tabular features into a single machine learning system. Such multimodal machine learning systems~\cite{baltruvsaitis2018multimodal} have been gaining popularity in recent years due to the availability of different data types such as texts, tabular data, images, sounds, etc., for each application. Previous works also combined multimodal machine learning and AutoML~\cite{nguyen2021multimodal, benchmarking} for text and tabular features. In~\cite{benchmarking}, several AutoML systems were compared on 18 different multimodal datasets that contain both texts and hand-crafted features. In~\cite{nguyen2021multimodal}, AutoML was used for credit modeling on financial data that combines raw texts from Securities and Exchange Commission filings with features extracted by financial experts. In these works, multimodal AutoML can achieve high accuracies while minimizing the implementation efforts. In this paper, we show that multimodal AutoML will also be effective for real estate data in a low-resource language (i.e., Vietnamese).

\section{Fake Classified Advertisements Detection}
\label{sec:detection}

\begin{table}[!t]
\begin{center}
  \caption{Examples of a true advertisement (top) and a fake advertisement (bottom) in the original form in Vietnamese (after removing all HTML tags). The corresponding English translations are provided for reference. Texts in square brackets are added for readability.}
  \label{tab:examples}
  \vspace{-0.2cm}
  \resizebox{.88\textwidth}{!}{%
  \begin{tabular}{p{0.37\textwidth} p{0.5\textwidth} c}
    \toprule
    Original advertisement & English translation & Label \\
    \midrule
    \begin{otherlanguage}{vietnamese} \vspace{-0.25cm} Chính chủ cần bán căn mặt tiền đường An Bình, Quận 5. DT: 3.8x16m. Nhà trệt, 3 lầu sân thượng. Vị trí: Khu chuyên kinh doanh vật phẩm phong thủy, kế chợ Hòa Bình, nhà hàng Đồng Khánh, Trần Hưng Đạo... Cần bán giá 15 tỷ. LH 0911142121. \end{otherlanguage}
    &
    The owner needs to sell a house on the main road of An Binh Street, District 5. [The house] area is 3.8$\times$16m, [with a] ground floor, 3 stories, [and a] rooftop terrace. [Its] location [is] in a special zone for shops selling Feng Shui products, next to Hoa Binh Market, Dong Khanh restaurant, Tran Hung Dao Street, etc. Need to sell for 15 billion [VND]. Contact 0911142121. & Real \\
    \midrule
    \begin{otherlanguage}{vietnamese} \vspace{-0.25cm} Nhà MT An Bình, Phường 6, Quận 5: DTKV: 10x40m, công nhận 400m2, nhà cấp IV, đang để trống, cần bán giá 55.5 tỷ. Đơn giá chỉ 137 triệu/m2, đảm bảo không còn sản phẩm so sánh. Vị trí mặt tiền thuận tiện kinh doanh, khuôn viên lớn phù hợp với nhiều ngành nghề hoặc xây cao cấp, building. Liên hệ 0906681528 Quang Dương để xem BĐS trên. Trân trọng cảm ơn quý khách. \end{otherlanguage}
    & The house is on the main road of An Binh Street, Ward 6, District 5. [The] total area [is] 10$\times$40m, [with an] official area [of] 400m$^2$. [This is a] bungalow and currently vacant. [The owner] needs to sell for 55.5 billion [VND]. [The] unit price is just 137 million/m$^2$. [We] guarantee this is the best price. The location on the main road is convenient for business. Its large area is suitable for various purposes or constructing luxurious buildings. Contact 0906681528 Quang Duong for viewing. Thank you sincerely.
    & Fake \\
    \bottomrule
  \end{tabular}
  }
\end{center}
\end{table}

We consider the problem of detecting fake advertisements among a collection of online classified advertisements, each of which is given as a free-form (unstructured) text. As illustrations, Table~\ref{tab:examples} shows two classified advertisements in their original form, where the first is a real advertisement and the second is a fake one. The original form of these advertisements is a free-form and unstructured text that is collected from an HTML \texttt{<textarea>} element with all the HTML tags removed. These texts usually contain a wide range of characters (alphabet letters, numbers, special symbols, etc.) and may include several paragraphs.

Detecting fake advertisements from these free-form texts is a challenging problem, especially in the context of Vietnamese classified real estate ads. The followings are some of the main reasons that make this problem difficult:

\begin{enumerate}
    \item \emph{The texts do not have a structure.}
    The ads do not have any pre-defined structure. Their contents depend totally on the users who post the ads. For instance, the sentences in the ads often do not follow proper grammatical rules, making it difficult to infer their semantics. In other cases, a piece of information can be provided in different ways (e.g., an ad can provide the length and width of a real estate instead of its total area), and we need a method to unify these information. These problems are true for both real and fake ads. 
    
    To illustrate these issues, we provide two examples of fake advertisements. One is the second advertisement in Table~\ref{tab:examples} and the other is the example in Table~\ref{tab:comp1_example}. We can easily notice from these advertisements that they have no general structures. For instance, the real estate agent can choose to describe the advantages of buying the property instead of describing the nearby facilities. They may also use different ways to provide the number of floors (bungalow, two floors, etc.).

    \item \emph{The texts contain a lot of abbreviations and redundant information.}
    Both true and fake ads tend to use abbreviations to shorten street names or some properties of the real estates (e.g., the ads in Table~\ref{tab:examples} use the abbreviations ``DT'' and ``DTKV'' for ``area''). There are no universal rules for the abbreviations. Furthermore, many ads also contain redundant information (such as information about nearby real estates or more than two properties in one listing). These types of ambiguity pose a challenge for automated learning systems, which need to disambiguate these information based on the context. These issues are illustrated in the first example of Table~\ref{tab:difficulties} where both abbreviations and redundancy are present.
    
    \item \emph{The texts lack important information.}
    The ads may not provide all the important information that the buyers need. This problem occurs not only in the fake ads but also in the real ads. In these cases, the owners or real estate agents can exclude some crucial information about the properties to attract prospective buyers. This practice could help increase the viewing rates for the properties and for the owners to better negotiate with prospective buyers. A sample fake advertisement that lacks the necessary information is given in the second example of Table~\ref{tab:difficulties}.

    \item \emph{Sparse data.}
    One of the hardest problems that we have to deal with is sparse data. In most cases, there are very few advertisements in a specific location. For example, on many streets, there could be less than five advertisements. This issue forces the models to make predictions on properties that are very different from those in the training data.
\end{enumerate}

\begin{table}[!t]
\begin{center}
  \caption{Examples of two fake advertisements that pose some difficulties mentioned in Section~\ref{sec:detection}. The corresponding English translations are provided for reference. Texts in square brackets are added for readability. The third column explains the difficulties that automatic detectors may encounter on these ads.}
  \label{tab:difficulties}
  \vspace{-0.2cm}
  \resizebox{.88\textwidth}{!}{%
  \begin{tabular}{p{0.26\textwidth} p{0.37\textwidth} p{0.3\textwidth}}
    \toprule
    Original advertisement & English translation & Difficulties for detectors \\
    (after preprocessing) & & \\
    \midrule
    \begin{otherlanguage}{vietnamese} 
    \vspace{-0.25cm} hxh 8m nguyễn trãi, q5 4,2mx16m, 2 tầng cực kỳ an ninh vip giá chỉ hơn 10 tỷ. hẻm nội bộ khu dân trí cao có hẻm phụ sau nhà 2m, 3pn gần mặt tiền đường nguyễn trãi xe hơi vô nhà, nhà mới 80\% nội thất cao cấp, nay cần tiền bán gấp giá 10.5 tỷ có thương lượng. và 1 căn hẻm xe hơi 8m an dương vương tuyến đường kinh doanh phụ tùng xe hơi lớn nhất thành phố thông ra trần bình trọng gần trường đại học sài gòn diện tích: 5mx21,6m, trệt 4 lầu giá 23.5 tỷ có thương lượng. liên hệ: 0914 194 586 gặp chí bảo. \end{otherlanguage}
    &
    [The house is located at an] 8-meter alley [on] Nguyen Trai [Street], District 5, [the area is] 4.2m$\times$16m, 2 floors, high security with price only over 10 billion [VND]. The neighborhood is highly educated with a 2-meter alley at the back, 3 bedrooms, near the front of Nguyen Trai Street. [The house can] fit a car, new house [with] 80\% high quality furniture. [We] need to sell urgently for 10.5 billion [or a] negotiable [price]. And [a house is located] at [an] 8-meter alley [on] An Duong Vuong [Street], a road [with the most] auto part businesses in the city [and connected with] Tran Binh Trong [Street], near Sai Gon University. [The area is] 5m$\times$21.6m, 4 floors [and a] ground floor, [the price is] 23.5 billion, negotiable. Contact 0914 194 586 and talk to Chi Bao.
    & 
    $\bullet$ {\bf Many abbreviations}: This advertisement contains many abbreviations such as hxh, q5, 3pn, etc. Here hxh means a kind of alley whose width is larger than 5 meters. This word can be inferred for the two types of features (house\_type and road\_width) mentioned in Table~\ref{tab:features}.

    $\bullet$ {\bf Redundant information}: This advertisement contains information about two different properties, one located on Nguyen Trai Street and the other on An Duong Vuong Street. The redundant information poses a challenge for detection systems to disambiguate the information.
    \\
    \midrule
    \begin{otherlanguage}{vietnamese} 
    \vspace{-0.25cm} kẹt tiền kinh doanh cần bán gấp căn nhà. dt: 4x18m nở hậu 4.2m đẹp. hiện trạng: trệt, 3 lầu. hẻm xe tải chạy thoải mái, khu sạch sẽ, an ninh, dân trí cao, gần trường học, bệnh viện lớn... gần hồng bàng, nguyễn chí thanh, châu văn liêm... hợp mua ở, cho thuê, làm phòng mạch, căn hộ dịch vụ. sổ hồng chính chủ. lh 0911142121.
    \end{otherlanguage}
    &
    [Due to the] shortage of money for our business, [we are] selling our house urgently. The area [is] 4$\times$18m with 4.2-meter wider at the back. The current condition is 3 floors and a ground floor. The alley is wide, clean, safe, well educated, and near to schools and hospitals, etc. Near Hong Bang Street, Nguyen Chi Thanh Street, Chau Van Liem Street, etc. [The house is] suitable for living, renting, and using as an office or service house. The ownership certificate is authentic. Contact 0911142121.
    &
    $\bullet$ {\bf Missing information}: This advertisement does not contain some crucial information about the property such as the price and location. Instead of the exact location, the advertisement only gives general information about the nearby facilities and advantages of the property. From the text, it is challenging to infer the exact road name and road features of the property.
    \\
    \bottomrule
  \end{tabular}
  }
\end{center}
\end{table}

\section{FADAML: Fake Advertisements Detection using Automated Multimodal Learning}
\label{sec:method}

In this paper, we propose a novel end-to-end system to tackle the fake classified advertisements detection problem above. Our approach, called \emph{Fake Advertisements Detection using Automated Multimodal Learning (FADAML)}, leverages recent advances in AutoML and multimodal machine learning to process the data and train machine learning models that can effectively detect fake advertisements from raw and unstructured texts. By combining automated feature extraction in an AutoML system and carefully hand-crafted multimodal features, our FADAML approach can overcome the difficulties mentioned above and achieve a good performance for this problem.

\begin{figure}[!t]
  \centering
  \includegraphics[width=.88\textwidth]{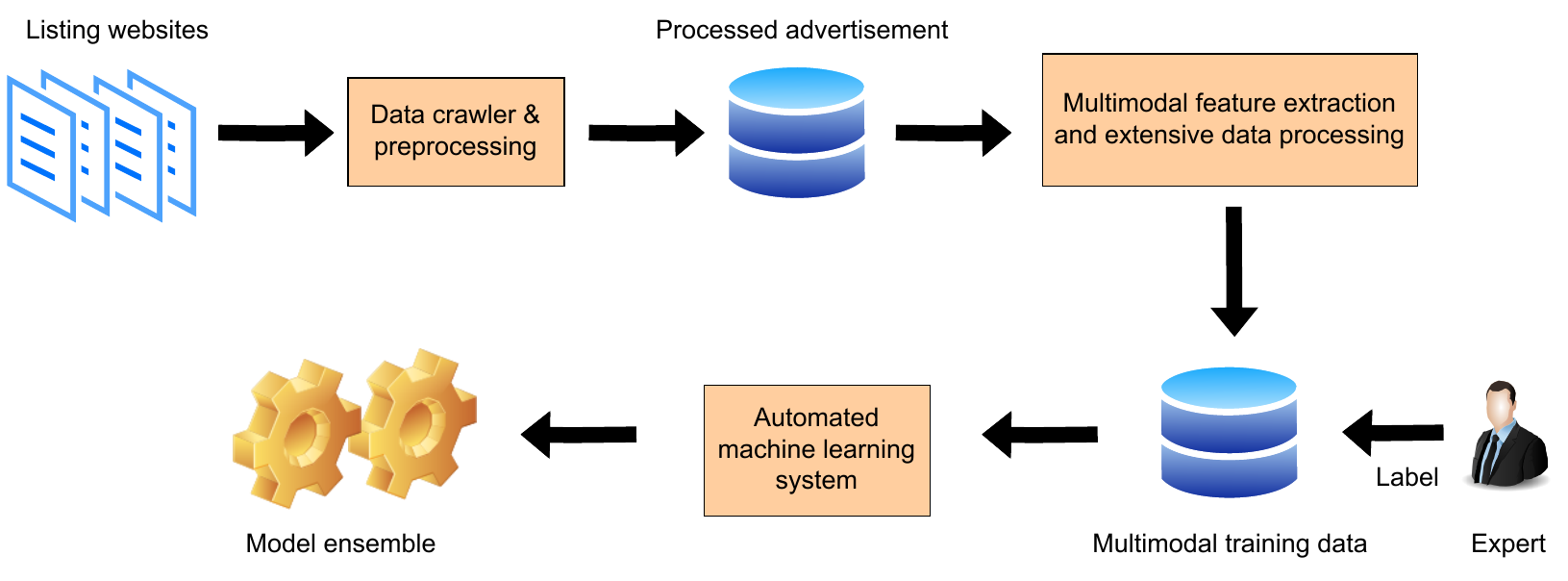}
  \vspace{-0.2cm}
  \caption{Components of FADAML, our proposed end-to-end system for detecting fake classified online advertisements using automated and multimodal machine learning.}
  \label{fig:system}
\end{figure}

The components of our system are visualized in Figure~\ref{fig:system}. The system consists of three main components: (1) a data crawler and preprocessing component, (2) a data processing and multimodal feature extraction component, and (3) an AutoML system. Below we describe these components in detail.

\subsection{Component 1: Data Crawler and Preprocessing}
\label{sec:com1}

In this first component, we develop a data crawler to collect data from a pre-defined list of domains. For each domain, our crawler automatically browses the webpages containing the real estate listings, downloads the corresponding HTML page, and extracts the real estate description from the page using its HTML structure. After collecting the descriptions of real estate posts from these webpages, we perform a light preprocessing step on these descriptions that includes: cleaning HTML elements such as \verb!\n!, \verb!\t!, \verb!<\br>!, etc.~using regular expressions, removing redundant white spaces, and converting the texts to lowercase.

In the context of Vietnamese classified real estate ads, we crawl raw data from the five most popular real estate advertisement domains: \textit{batdongsan.com.vn}, \textit{chotot.com}, \textit{diaoconline.vn}, \textit{123nhadat.vn}, and \textit{muaban.net}. These raw advertisements are then processed by our preprocessing modules described above. As an illustration, we show in Table~\ref{tab:comp1_example} an example of a raw advertisement returned by the crawler and the corresponding preprocessed advertisement. This preprocessed advertisement will be the input to the second component of our FADAML system, which we will describe next.

\begin{table}[!t]
\begin{center}
  \caption{An example showing the outputs of the first component of FADAML, which consists of a data crawler and a preprocessing module. The left column shows a raw advertisement returned by the crawler, the center column shows the preprocessed advertisement, and the right column shows the English translation of the raw advertisement.}
  \label{tab:comp1_example}
  \vspace{-0.2cm}
  \resizebox{.88\textwidth}{!}{%
  \begin{tabular}{p{0.32\textwidth} p{0.28\textwidth} p{0.36\textwidth}}
    \toprule
    Raw advertisement & Advertisement after the & English translation \\
    returned by crawler & preprocessing step & \\
    \midrule
    \begin{otherlanguage}{vietnamese} \vspace{-0.2cm} Bán nhà mặt tiền đường Nguyễn Chí Thanh, phường 12, quận 5. Đoạn ngay ngô quyền và nguyễn chí thanh. Gần bệnh viện chợ rẫy, <br/> (10x28m) nhà nở hậu 11m. Diện tích công nhận: 290m2. Nhà 2 lầu. <br/>DT: 10x30m, nở hậu nhẹ, CN 290m2.<br/>Tiện xây mới tòa nhà vp, ngân hàng, khách sạn,... Khu nhiều bệnh viện, trường học, đầy đủ tiện ích xung quanh. <br/>Giá bán: 60.5 tỷ thương lượng lấy lộc với khách thiện chí.<br/>Quý khách mua và xem nhà liên hệ Trần Toàn: 0918 462 562. <br/>- Thông tin chính xác 100\%. Người thật, việc thật. \end{otherlanguage}
    &\begin{otherlanguage}{vietnamese} \vspace{-0.2cm}
    bán nhà mặt tiền đường nguyễn chí thanh, phường 12, quận 5, đoạn ngay ngô quyền và nguyễn chí thanh, gần bệnh viện chợ rẫy. (10x28m) nhà nở hậu 11m. diện tích công nhận: 290m2, nhà 2 lầu, dt: 10x30m, nở hậu nhẹ, cn 290m2. tiện xây mới tòa nhà vp, ngân hàng, khách sạn,... khu nhiều bệnh viện, trường học đầy đủ tiện tích xung quanh. Giá bán: 60.5 tỷ thương lượng lấy lộc với khách thiện chí. quý khách mua và xem nhà liên hệ trần toàn: 0918 462 562.- thông tin chính xác 100\% người thật, việc thật.  
    \end{otherlanguage}
    & Selling a house on the main road of Nguyen Chi Thanh Street, ward 12, district 5, between Ngo Quyen and Nguyen Chi Thanh streets, near Cho Ray hospital. (10x28m) The rear [of the plot is] 11m wider [than the front]. Official area: 290m2, with 2 floors, total area: 10x30m, slightly wider rear, official [area] 290m2. Convenient for building offices, banks, hotels, etc. The zone has many hospitals, schools, and all nearby facilities. Selling price: 60.5 billion, [willing to] negotiate with customers with goodwill. Customers [who want] to buy and see the house, [please] contact Tran Toan: 0918 462 562. - 100\% accurate information, real person, real listing. \\
    \bottomrule
  \end{tabular}
  }
\end{center}
\end{table}

\subsection{Component 2: Multimodal Feature Extraction and Extensive Data Processing}
\label{sec:com2}

\begin{figure}[!t]
  \centering
  \includegraphics[width=.84\textwidth]{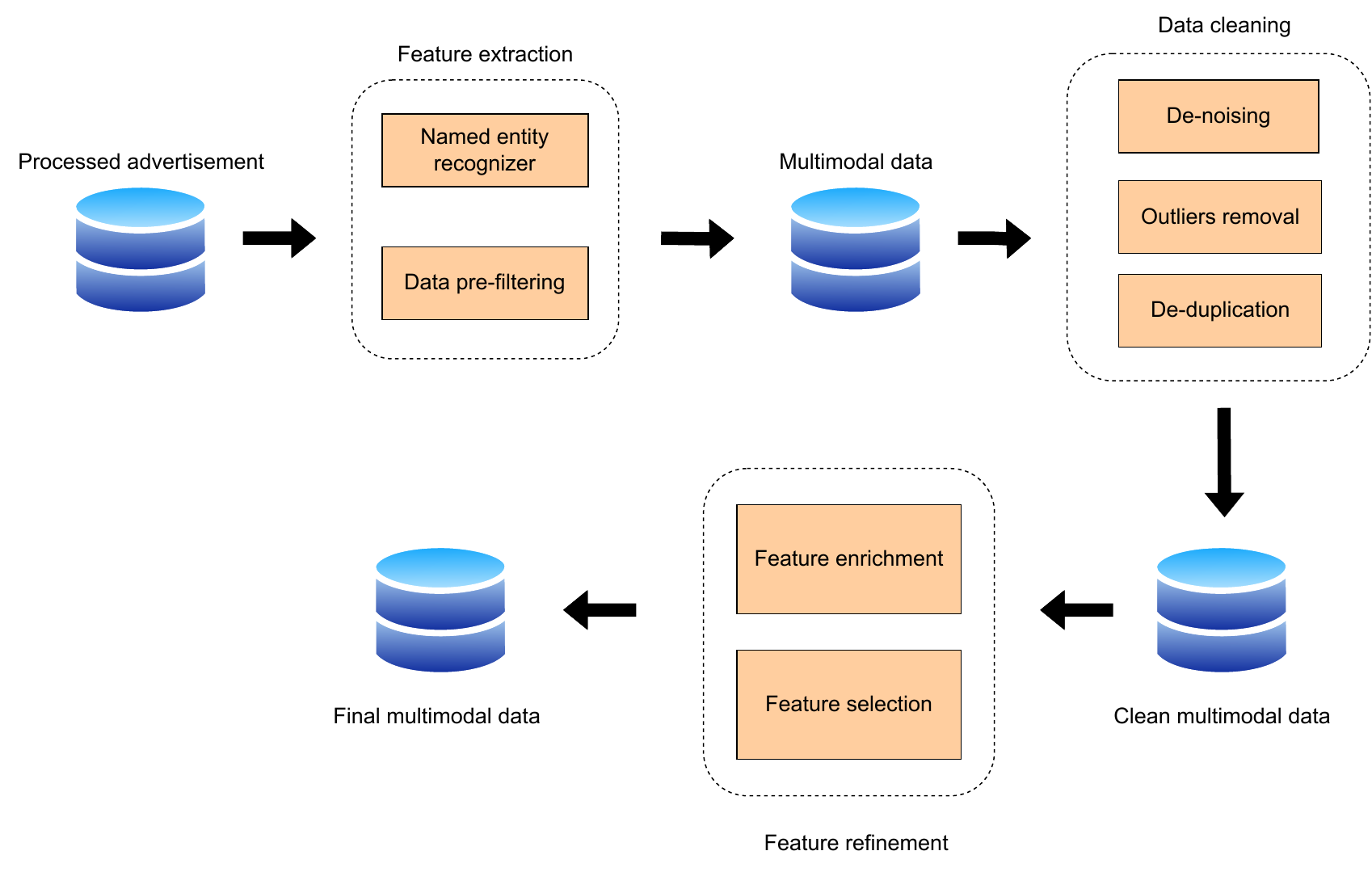}
  \vspace{-0.2cm}
  \caption{The second component of our FADAML system. This component includes three processing steps: multimodal feature extraction, data cleaning, and feature refinement.}
  \label{fig:processing}
\end{figure}

The second component of our FADAML system takes the processed advertisements from the first component as inputs and applies the following three processing steps: (1) multimodal feature extraction, (2) data cleaning, and (3) feature refinement. An illustration of this component is presented in Figure~\ref{fig:processing}. We detail each of the processing steps below.

\subsubsection{Multimodal Feature Extraction}
\label{sec:feat-extract}

The main purpose of this step is to parse and extract a set of tabular features from the previously preprocessed advertisements that will be combined with the advertisement text itself to form a set of multimodal features. The advertisement text will later be automatically processed by the AutoML component in Section~\ref{sec:automl}. To extract the tabular features, we use a named entity recognizer pre-trained on a different dataset to extract a set of named entities useful for our problem. In particular, we apply the named entity recognizer proposed by Huynh et al.~\cite{ner} that has been trained separately on another real estate dataset collected and annotated with Doccano~\cite{doccano}. This recognizer uses the MishWindowEncoder model that takes an embedding vector from PhoBERT~\cite{nguyen2020phobert} and returns the corresponding feature vector. This feature vector is then fed to a classifier to be classified into the target named entities. Using this recognizer, we can extract four most useful entities for our problem: \emph{price}, \emph{area}, \emph{road}, and \emph{district} (see Table~\ref{tab:features} for the descriptions of these features).

After extracting the named entities and the tabular features, we then perform a data pre-filtering step to remove all advertisements not related to real estates or those that do not contain sufficient information for a transaction (e.g., those without information on price, location, or property area). This step is performed by another classifier trained separately on a dataset customized for this task. This classifier will determine whether a listing is about some real estate transaction and does not contain inappropriate or offensive content.

\subsubsection{Data Cleaning}

After constructing the multimodal features from the previous step, we perform extensive data cleaning in this step to ensure all data records are clean and normalized. From our preliminary inspection, we notice there are three main issues with the data, namely noisy records, outliers, and duplicated posts. Thus, we perform the following three data cleaning tasks to address these issues.

\emph{De-noising.} In the context of our paper, we define a data record as noisy if the advertisement does not provide reliable or crucial information about a property, or if the advertisement is too complicated to extract useful features. These noisy data records often contain irrelevant information that is not useful for the intended real estate advertisement. For instance, a listing may contain more than two different locations or addresses for one property or there is a discrepancy between the street and the district of the property. To detect these noisy records, we analyze the extracted features and employ a set of heuristic rules based on our domain knowledge (e.g., missing features) to help us detect and drop these records.

\emph{Outlier removal.} In addition to noisy data records, we also detect and remove outliers, which are defined as records containing abnormal feature values. For instance, on a website's section for selling properties, renting advertisements are considered outliers since their extracted price is much lower than the usual selling price. Another example is the incorrectly extracted features from a very complex raw description text. In this paper, we detect the outliers using pre-defined ranges for each feature values in consultation with real estate experts. Depending on the type of the errors, we will decide to either drop or fix the data records.

\emph{De-duplication.} Since our data are collected from multiple listing websites, the records can be duplicated due to the cross-posting of real estate agencies to increase the advertisements' visibility. In most cases, these duplicated postings contain the same information but are written using different formats and structures, or even with some added redundancies. In this task, we implement a heuristic de-duplication process to filter out these duplicated records. First, we de-duplicate by dropping records with similar description texts. Then we compare the important features and de-duplicate the records containing similar information. This process helps us significantly reduce the number of duplicated records in practice.

\subsubsection{Feature Refinement}
\label{label:feat-refine}

This step is to further refine the features before training. In this step, we perform feature selection and feature enrichment on the multimodal data obtained from the previous step. Since there are several features that do not have any impact on the value of a real estate, the feature selection step chooses a subset of the named entities obtained from the previous feature extraction step to be our features. In particular, from the experts' opinions, we know that fake real estate advertisements can often be detected by inspecting the discrepancy between the content of the ad and the posted property price. Thus, we first sort the features using their correlation scores with the property's price and then select a subset of the highly scored features that overlap with the experts' frequently used features. During this feature selection step, we also maintain the consistency between all records such that they would have the same value type and unit on the same features (e.g., all enumerated district names are converted to strings).

Working with real estate experts, we further perform feature enrichment to enhance the prediction performance. The feature enrichment process allows us to extract new features that cannot be recognized by the named entity recognizer in the previous step. In Table~\ref{tab:features}, we give details of all features used in our system, including those extracted from the previous feature extraction step as well as those from the feature enrichment step.

\begin{table}[!t]
\begin{center}
  \caption{The list of all features used in our FADAML system together with their types and descriptions. The last column indicates which step in our pipeline generates the feature.}
  \label{tab:features}
  \vspace{-0.2cm}
  \resizebox{.85\textwidth}{!}{%
  \begin{tabular}{p{0.12\textwidth} p{0.1\textwidth} p{0.39\textwidth} p{0.25\textwidth}}
    \toprule
    Feature & Type & Description & Generated from\\
    \midrule
    description & string & The preprocessed advertisement text. & Component 1 \newline (Section~\ref{sec:com1}) \\[5pt]
    price & real & The total price of the real estate (in million Vietnamese Dongs). & Multimodal feature \newline extraction (Section~\ref{sec:feat-extract}) \\[5pt]
    area & real & The ground area of the real estate (in square meters). & Multimodal feature \newline extraction (Section~\ref{sec:feat-extract}) \\[5pt]
    house\_type & category & A category indicating whether the real estate is located on the main road (frontage house) or in an alley (alley house). & Feature enrichment \newline (Section~\ref{label:feat-refine}) \\[5pt]
    road & category & The name of the road where the real estate is located. & Multimodal feature \newline extraction (Section~\ref{sec:feat-extract}) \\[5pt]
    district & category & The name of the district where the real estate is located. & Multimodal feature \newline extraction (Section~\ref{sec:feat-extract}) \\[5pt]
    road\_width & real & The width of the road where the alley house is located. If the real estate is on a main road, this feature is set to a fixed number. & Feature enrichment \newline (Section~\ref{label:feat-refine}) \\[5pt]
    road\_first & category & The nearest road in the same district with the real estate. & Feature enrichment \newline (Section~\ref{label:feat-refine}) \\[5pt]
    road\_second & category & The second nearest road in the same district with the real estate. & Feature enrichment \newline (Section~\ref{label:feat-refine}) \\[5pt]
    road\_third & category & The third nearest road in the same district with the real estate. & Feature enrichment \newline (Section~\ref{label:feat-refine}) \\[5pt]
    \bottomrule
  \end{tabular}
  }
\end{center}
\end{table}

Among the features, we use both the real estate description texts obtained from Component 1 and the features extracted in Component 2. The \emph{price}, \emph{area}, \emph{road}, and \emph{district} features are obtained from the feature extraction step and are standardized to ensure all data samples have the same data type for these features. The \emph{house\_type} and \emph{road\_width} features are generated from the feature enrichment process, where we use regular expressions to capture them. If the real estate is located on a main road, we set \emph{road\_width} to be a large constant number (we use 20 in this case). If the real estate is an alley house and there is no information about the road width in the description, we set \emph{road\_width} to a fixed constant based on the location of the house. The features added during the above process and their values are obtained by consulting a team of two real estate experts, who would need to discuss and completely agree on the final suggestions.

To obtain the \emph{road\_first}, \emph{road\_second}, and \emph{road\_third} features, we use the GeoPy package\footnote{\url{https://geopy.readthedocs.io/}.} to get the longitude and latitude of each road in the district where the real estate is located. With these longitudes and latitudes, we calculate the relative distance between each pair of roads by using Manhattan distance \cite{krause1986taxicab}. We sort the relative distances to get the top three nearest roads and use them as features. These features allow us to take into account the spatial information of the roads, where nearby roads within the same district would likely have similar prices per unit area. Another advantage of using these nearest road features is their ability to deal with missing data. In many practical situations, there could be too few or even no advertisements at a specific location in the same time, thus the machine learning models may need to deal with properties on a new road during testing. In this case, the nearest road features can be used to extrapolate the information for the main real estate.

It is worth noting that the multimodal features extracted in this component can help our system handle sparse data effectively. In particular, features such as \emph{price}, \emph{area}, \emph{house\_type}, \emph{district}, \emph{road\_width}, \emph{road\_first}, \emph{road\_second}, and \emph{road\_third} can give us a lot of information about a property even if it is the only property on a given road. This intuition is confirmed by the importance scores of these features in our experiment results in Section~\ref{sec:experiments}.

\begin{table}[!t]
\begin{center}
  \caption{A final data sample that will be input to our AutoML system, which will convert the description string into generic natural language features for training and prediction.}
  \label{tab:processed}
  \vspace{-0.2cm}
  \resizebox{.86\textwidth}{!}{%
  \begin{tabular}{p{0.2\textwidth} p{0.7\textwidth}}
  \toprule
  Feature & Value \\
  \midrule
  description & \begin{otherlanguage}{vietnamese} \vspace{-0.25cm} nhà mt an bình, phường 6, quận 5: dtkv: 10x40m, công nhận 400m2, nhà cấp iv, đang để trống, cần bán giá 55.5 tỷ. đơn giá chỉ 137 triệu/m2, đảm bảo không còn sản phẩm so sánh. vị trí mặt tiền thuận tiện kinh doanh, khuôn viên lớn phù hợp với nhiều ngành nghề hoặc xây cao cấp, building. liên hệ 0906681528 quang dương để xem bđs trên. trân trọng cảm ơn quý khách.\end{otherlanguage} \\[5pt]
  price & 55,500 \\[5pt]
  area & 400 \\[5pt]
  house\_type & \begin{otherlanguage}{vietnamese} \vspace{-0.25cm} nhà mặt tiền \end{otherlanguage} (frontage house) \\[5pt]
  road & \begin{otherlanguage}{vietnamese} \vspace{-0.25cm} an bình \end{otherlanguage} \\[5pt]
  district & 5 \\[5pt]
  road\_width & 20 \\[5pt]
  road\_first & \begin{otherlanguage}{vietnamese} \vspace{-0.25cm} trần phú \end{otherlanguage} \\[5pt]
  road\_second & \begin{otherlanguage}{vietnamese} \vspace{-0.25cm} trần tuấn khải \end{otherlanguage} \\[5pt]
  road\_third & \begin{otherlanguage}{vietnamese} \vspace{-0.25cm} võ văn kiệt \end{otherlanguage} \\[5pt]
  \bottomrule
  \end{tabular}
  }
\end{center}
\end{table}

\subsection{Component 3: Automated Machine Learning System}
\label{sec:automl}

After executing the first two components of our FADAML system, we obtain the full multimodal dataset with 10 features in Table~\ref{tab:features}. We show in Table~\ref{tab:processed} a final data sample that will be input to our third component, the AutoML system, which will automatically perform all standard feature processing techniques that are known to improve the predictive performance of the trained model.

To obtain the ground truth labels, we employ an approach involving both human and machine. In consultation with real estate experts, we know that fake Vietnamese real estate ads often have a large discrepancy between the posted price and the evaluated price based on the ad's content. Thus, we obtain the ground truths by inspecting the posted price of each property and its predicted price based on past transactions of similar properties. When there is a large discrepancy between the prices, two experts will be asked to independently assess the ad's content and give their evaluated prices. If the prices from the two experts are not within 10\% of each other, the ad is considered fake. After assigning the labels, the experts will manually check and correct a random subset of labels. This process is repeated until no correction is made.

After receiving the labeled dataset, our AutoML system will first perform an encoding step to convert categorical features into numeric and generate several basic natural language features from the text descriptions (i.e., from the first feature in Table~\ref{tab:features}). The natural language features include n-gram counts, character counts, word counts, special symbol counts, special symbol ratios, and digit ratios. The categorical features are converted into numeric using a label encoder, which uses monotonically increasing integer values to reduce memory usage.

Following previous works that utilized AutoML~\cite{nguyen2021multimodal, agtabular}, we also use model stacking to achieve the best performance. Particularly, we use an ensemble of several base models stacked into two layers automatically using the validation set. The list of base models trained by our system is described in Table~\ref{tab:describe_model}. To obtain the 2-layer stack ensemble, we first train all the base models with the training set that would constitute the first layer of the stack. To obtain the second layer, we concatenate the predictions of the first layers with the original features to form a new input vector, and then apply ensemble selection~\cite{ensemble} to construct the second layer. This ensemble selection step will iteratively add to the ensemble the model that most improves the performance on a validation set. 

By considering the accuracy of each model, we can also aggregate the previous layer outputs into a weighted ensemble. To make a prediction, the ensemble will calculate the weighted average of the predictions from its base models. We compare the predictions of the ensemble and the base models on the validation dataset to get the best configuration. After multiple trials, we observe that the weighted ensemble yields the best accuracy in most cases.

\begin{table}[!t]
\begin{center}
  \caption{The base models used in our AutoML system.}
  \label{tab:describe_model}
  \vspace{-0.2cm}
  \resizebox{.85\textwidth}{!}{%
  \begin{tabular}{p{0.25\textwidth} p{0.68\textwidth}}
  \toprule
  Model & Description \\
  \midrule
  Categorical Boosting \newline (CatBoost)~\cite{catboost} & Catboost is a powerful gradient boosting algorithm that can handle categorical features and uses the ordered boosting technique to prevent prediction shift and data leakage. Catboost is experimentally proven to outperform previous gradient boosting algorithms on various machine learning tasks. \\[5pt]

  Extreme Gradient \newline Boosting (XGBoost)~\cite{xgboost} & XGBoost is a scalable end-to-end boosting algorithm that can often provide state-of-the-art accuracy on tabular data. The model uses a sparsity-aware algorithm for parallel tree learning and a weighted quantile sketch for approximate tree learning. \\[5pt]

  LightGBM~\cite{lightgbm} & LightGBM is a popular gradient boosting decision tree algorithm that is very efficient and scalable for high-dimensional data and large data sets. LightGBM uses gradient-based one-side sampling and exclusive feature bundling techniques to exclude a significant proportion of data with small gradients and bundle mutually exclusive features to obtain a decent estimation of the information gain and speed up the training process. \\[5pt]

  Random Forest~\cite{breiman2001random} & Random forest is an ensemble machine learning algorithm that trains a model by constructing multiple decision trees at training time. It is an extension of the bagging method that utilizes both bagging and feature randomness to create an uncorrelated forest of decision trees. \\[5pt]
  
  Extra Tree \newline Classifier~\cite{geurts2006extremely} & This classifier implements the extremely randomized tree algorithm. Instead of using a random subset of candidate features, this algorithm uses the best randomly generated thresholds for each candidate feature to split the trees. It helps reduce the variance of the model but also slightly increases the bias. \\[5pt]

  Neural network~\cite{bishop1994neural} & We use the multilayer perceptron model, which comprises an input layer, one or more hidden layers, and an output layer. This is a basic deep learning model that has been successfully applied to several applications. \\[5pt]

  K-nearest \newline neighbors~\cite{dudani1976distance} & This is a classic non-parametric supervised learning algorithm that uses proximity to a group of neighboring points and their labels to make predictions. \\[5pt]
  \bottomrule
  \end{tabular}
  }
\end{center}
\end{table}

\section{Experiments}
\label{sec:experiments}

In this section, we empirically evaluate the performance of our FADAML system on real-world Vietnamese real estate data. We will first describe our experiment settings in Section~\ref{sec:experiment-setting} and then our results in Section~\ref{sec:exp-result}. We also give an ablation study for our system in Section~\ref{sec:ablation} and discuss some limitations of our work in Section~\ref{sec:discussion}. Our experiments are conducted on a computer with an AMD Ryzen 7 (3.20GHz) processor and 16GB of RAM.

\subsection{Experiment Settings} 
\label{sec:experiment-setting}

\subsubsection{Settings}

\begin{table}[t!]
\begin{center}
  \caption{Statistics of training and testing datasets.}
  \label{tab:data}
  \vspace{-0.2cm}
  \small{
  \begin{tabular}{lccc}
    \toprule
    Information & Train & Test & All \\
    \midrule
    Number of posts/examples & 23,268 & 5,817 & 29,085\\
    Number of real posts/examples & 9,788 & 2,447 & 12,235\\
    Number of fake posts/examples & 13,480 & 3,370 & 16,850\\
    \bottomrule
    \end{tabular}
    }
    \vspace{-0.2cm}
\end{center}
\end{table}

We use the process described in Section~\ref{sec:com1} to collect and preprocess the data for our experiment. Then we use the second component of our system in Section~\ref{sec:com2} to extract the multimodal features as well as refining the data. After these steps, we obtain a dataset of 29,085 examples, which we randomly split into a training and a testing set. The statistics of these datasets are in Table~\ref{tab:data}. During training, we also use stratified random sampling to select a validation set from the training set. The validation set constitutes 10\% of the original training set.

In our named entity recognition step in Section~\ref{sec:feat-extract}, we use PhoBERT~\cite{nguyen2020phobert} to obtain the embeddings of our text inputs. PhoBERT is a public large-scale language model pre-trained for the Vietnamese language that has achieved state-of-the-art performance on several natural language tasks. This model has two variants, PhoBERT$_{base}$ and PhoBERT$_{large}$, which use the same architectures of BERT$_{base}$ and BERT$_{large}$ respectively~\cite{devlin2019bert}. In our experiments, we use the PhoBERT$_{base}$ pre-trained model available on the HuggingFace repository~\cite{wolf2020transformers}.

We implement our automated machine learning system in Section~\ref{sec:automl} using AutoGluon 0.4.0~\cite{agtabular}, a robust open-source AutoML framework. We fix the n-gram range for our NLP features as (1, 3) and set the maximum number of features to be 10,000. Additionally, in order to reduce the memory usage during training, we set the maximum memory ratios at 0.15, i.e., we only use around 15\% of the total allocated memory for n-gram features and remove the least frequent n-grams. This method helps us avoid out-of-memory errors in the training phase.

We also take advantage of the AutoGluon pipeline to implement the 2-layer stack ensemble. Its training procedure is inspired by layer-wise training and skip connections in deep learning. The main difference here is that every node in an AutoGluon's layer represents a machine learning model. We do not set a time limit for training to allow AutoGluon to train all models until convergence. In Table~\ref{tab:model_setting}, we give the detailed settings used when training the AutoGluon models. After training these models, AutoGluon uses the validation set to obtain the best model. Since our experiment results show that the weighted ensemble is the best model, we use this ensemble model to make predictions on the test set.

\begin{table}[!t]
\begin{center}
  \caption{Training settings of the AutoGluon models used in our experiments.}
  \label{tab:model_setting}
  \vspace{-0.2cm}
  \resizebox{.86\textwidth}{!}{%
  \begin{tabular}{p{0.25\textwidth} p{0.65\textwidth}}
  \toprule
  Model & Training settings \\
  \midrule
  CatBoost & We train this model with learning rate 0.05 and set the maximum number of trees to be 10,000. \\[5pt]

  Extra Trees Entropy & This model implements the Extra Tree classifier with the entropy impurity. We set the maximum number of trees in the model to be 300. \\[5pt]

  Extra Trees Gini & This model implements the Extra Tree classifier with the Gini impurity. We set the maximum number of trees in the model to be 300. \\[5pt]

  Random Forest Entropy & This model implements the Random Forest classifier with the entropy impurity. We set the maximum number of trees in the model to be 300. \\[5pt]

  Random Forest Gini & This model implements the Random Forest classifier with the Gini impurity. We set the maximum number of trees in the model to be 300. \\[5pt]

  KNN Uniform & This model implements the K-nearest neighbors classifier with uniform weight for all data points in each neighborhood. \\[5pt]

  KNN Distance & This model implements the K-nearest neighbors classifier where each data point in a neighborhood is weighted by its inverse distance to the test point. \\[5pt]

  XGBoost & For this model, we use the gradient boosted tree (gbtree) booster and set the learning rate to 0.1. \\[5pt]

  LightGBM & This model implements the regular version of LightGBM. We use learning rate 0.05 to train the model. \\[5pt]

  LightGBM-Large & This model implements a version of LightGBM that has been customized for large datasets. We use learning rate 0.03 and set the maximum number of leaves in one tree to be 128. We only use 90\% of the features before training each tree to avoid overfitting and reduce the training time. \\[5pt]

  LightGBM-XT & This model implements a version of LightGBM that uses extremely randomized trees, where the model measures the quality of the node splits using only one threshold for each feature. This helps speed up training and avoid overfitting. We use learning rate 0.05 to train this model. \\[5pt]

  Neural Net Torch & This model implements the multi-layer perceptron classifier with 4 layers (128 nodes per layer) and ReLU activation using the PyTorch library~\cite{paszke2019pytorch}. We train this neural network for 500 epochs using the Adam optimizer~\cite{kingma2014adam} with learning rate $3 \times 10^{-4}$ and weight decay $10^{-6}$. \\[5pt]

  Neural Net FastAI & This model implements the default multi-layer perceptron classifier of the FastAI library~\cite{howard2020fastai}. We use learning rate 0.01, 10\% linear layer dropout, and early stopping to obtain the best results. \\[5pt]

  Weighted Ensemble & This model implements the 2-layer weighted ensemble of the above models. \\[5pt]
  \bottomrule
  \end{tabular}
  }
\end{center}
\end{table}

\subsubsection{Baselines}

In our experiments, we compare FADAML with the following three state-of-the-art systems for text classification and fake news detection. These baseline systems are chosen due to their effectiveness on problems similar to ours.

\begin{itemize}
  \item \textbf{FNDNet}~\cite{kaliyar2020fndnet}. This is a deep learning model that consists of three parallel convolutional layers operated on the word embedding vectors of the texts. To adapt this model to the Vietnamese language and our dataset, we use phoBERT~\cite{nguyen2020phobert} to generate the embedding vectors and manually tune each layer's kernel size to achieve the best possible performance.
  \item \textbf{Bi-LSTM+Attention}~\cite{antoun2020state}. This system uses ELMo~\cite{peters2018deep} to extract contextualised word embeddings from the texts, which is then fed to a Bidirectional Long Short Term Memory (Bi-LSTM) model with an attention layer to obtain the predictions. To adapt the method to our dataset, we use the ELMo embeddings for Vietnamese~\cite{che2018towards} while keeping the original architecture of the model.
  \item \textbf{FastText+CNN}~\cite{umer2023impact}. This is a state-of-the-art system that combines FastText word embeddings with convolutional neural network architecture to perform English text classification effectively. To use this system for our dataset, we utilize the Vietnamese FastText word embedding model with the original CNN model architecture.
\end{itemize}

\subsubsection{Evaluation Metrics}

To get a comprehensive view of the performance of the above systems, we evaluate and compare these systems using various metrics below. Note that in the following definitions, $T_p, T_n, F_p, F_n$ are the number of true positives, true negatives, false positives, and false negatives of a detection system, respectively.

\begin{itemize}
  \item \textbf{Precision}. This metric is defined as $T_p / (T_p + F_p)$, which measures the fraction of truly fake advertisements among those detected by a system.
  \item \textbf{Recall}. This metric is defined as $T_p / (T_p + F_n)$, which measures the fraction of correctly detected fake advertisements among all the fake ads.
  \item \textbf{F1}. This metric is defined as $2 P R / (P + R)$, where $P$ is the precision and $R$ is the recall of a system. The F1 metric measures the effectiveness of a detection system when precision and recall are given equal importance.
  \item \textbf{Accuracy}. This metric is defined as $(T_p + T_n) / (T_p + T_n + F_p + F_n)$, which measures the effectiveness of a detection system when the dataset is balanced. Since our dataset is only slightly imbalanced, comparing both the accuracy and F1 metrics gives us a good indicator of the effectiveness of the detection systems.
  \item \textbf{False positive rate (FPR)}. This metric is defined as $F_p / (F_p + T_n)$, which measures the false alarm rate where a legitimate ad is flagged as fake.
  \item \textbf{False negative rate (FNR)}. This metric is defined as $F_n / (F_n + T_p)$, which measures the rate that a system fails to detect fake advertisements.
\end{itemize}

\begin{table}[t]
\begin{center}
  \caption{Comparison of our FADAML system with state-of-the-art fake news detection and text classification systems on our dataset. The upwards arrows ($\uparrow$) indicate that higher numbers are better, while the downwards arrows ($\downarrow$) indicate that smaller numbers are better. Bold numbers indicate the best results in the corresponding column.}
  \label{tab:compare_result}
  \vspace{-0.2cm}
  \small{
  \begin{tabular}{lcccccc}
    \toprule
    Model & Precision$^\uparrow$ & Recall$^\uparrow$ & F1$^\uparrow$ & Accuracy$^\uparrow$ & FPR$^\downarrow$ & FNR$^\downarrow$ \\
    \midrule
    FNDNet~\cite{kaliyar2020fndnet} &0.726  &0.722  &  0.724& 0.733 &0.340 &0.213\\
    Bi-LSTM+Attention~\cite{antoun2020state} &  0.631 & 0.619 &0.620  & 0.643 & 0.529 & 0.231\\
    FastText+CNN~\cite{umer2023impact} & 0.289  & 0.500 & 0.366  & 0.579 & 1.00 & 0.00\\
    FADAML (ours) & {\bf 0.913} & {\bf 0.913} & {\bf 0.913} & {\bf 0.915} &{\bf 0.098}&{\bf 0.074}\\
    \bottomrule
  \end{tabular}
  }
\end{center}
\end{table}

\subsection{Results}
\label{sec:exp-result}

In Table~\ref{tab:compare_result}, we show the result comparing our FADAML system with the FNDNet, Bi-LSTM+Attention, and FastText+CNN baselines. From the table, we can see that FADAML outperforms the other systems significantly in all metrics. Specifically, FADAML achieves 91.3\% F1 score and 91.5\% accuracy while the second best system (FNDNet) only achieves 72.4\% F1 score and 73.3\% accuracy. Surprisingly, FastText+CNN, which is the most recent system~\cite{umer2023impact}, performs poorly on our dataset with 100\% false positive rate and 0\% false negative rate. This could be due to the more difficult vocabulary (e.g., many abbreviations in Vietnamese, real estate terminologies, listing structure, etc.) used in our dataset, making the system unable to learn meaningful patterns from the data.

The large difference in performance between our system and the baselines could be due to the addition of multimodal, refined, and spatial features in the FADAML system, as shown in our ablation study later. Another reason for the accuracy difference is that, when compared to the English datasets used in their original work, our Vietnamese dataset is more complicated for these baseline models. This result shows that FADAML is more effective for detecting Vietnamese fake advertisements than the other state-of-the-art systems. Our system is also more immune to false positives and false negatives compared to the baselines.

To better understand our FADAML system, we show in Table~\ref{tab:models} the training performance of the individual models returned by AutoGluon after training the system. From the table, the weighted ensemble achieves the best validation accuracy (92.5\%) among all models, followed by LightGBM-Large as the second best model (90.4\%) and then Neural Net Torch as the third best model (89.8\%). For our problem, the random forest, extra tree, KNN, and Neural Net FastAI models do not provide a good performance. The validation accuracies in this table clearly show that we should use the weighted ensemble as our final model for the best prediction accuracy. We note that our system uses around 15\% of the computer's memory (i.e., around 2.4GB) and the training time for the final system (i.e., the weighted ensemble model) from Table~\ref{tab:models} is around 124.86 seconds, which is reasonable given the accuracy of the system. However, if we want to further reduce the training time, we can restrict the number of base models in the ensemble with a possible accuracy trade-off.

\begin{table}[t]
\begin{center}
  \caption{Training performance of the individual models (sorted by descending order of accuracy) obtained from the best setting of our FADAML system. The validation accuracy and prediction time are measured on a validation split of the training data. Times are measured in seconds.}
  \label{tab:models}
  \vspace{-0.2cm}
  \small{
  \begin{tabular}{lccc}
    \toprule
    Model & Validation accuracy & Prediction time (s) & Training time (s) \\
    \midrule
    Weighted Ensemble & 0.925 & 0.376 & 124.860 \\
    LightGBM-Large & 0.904 & 0.063 & 6.696 \\
    Neural Net Torch & 0.898 & 0.032 & 25.454 \\
    LightGBM & 0.892 & 0.066 & 4.583 \\
    CatBoost & 0.892 & 0.208 & 72.13 \\
    XGBoost & 0.883 & 0.117 & 10.708 \\
    LightGBM-XT & 0.881 & 0.112 & 13.059 \\
    Random Forest Entropy & 0.790 & 0.066 & 14.881 \\
    Random Forest Gini & 0.787 & 0.066 & 14.958 \\
    Extra Trees Entropy & 0.782 & 0.070 & 19.921 \\
    Extra Trees Gini & 0.781 & 0.069 & 19.680 \\
    KNN Distance & 0.756 & 10.487 & 1.131 \\
    KNN Uniform & 0.721 & 10.681 & 1.101 \\
    Neural Net FastAI & 0.627 & 0.027 & 9.158 \\
    \bottomrule
  \end{tabular}
  }
\end{center}
\end{table}

\begin{figure}[h]
\centering
\includegraphics[width=0.6\textwidth]{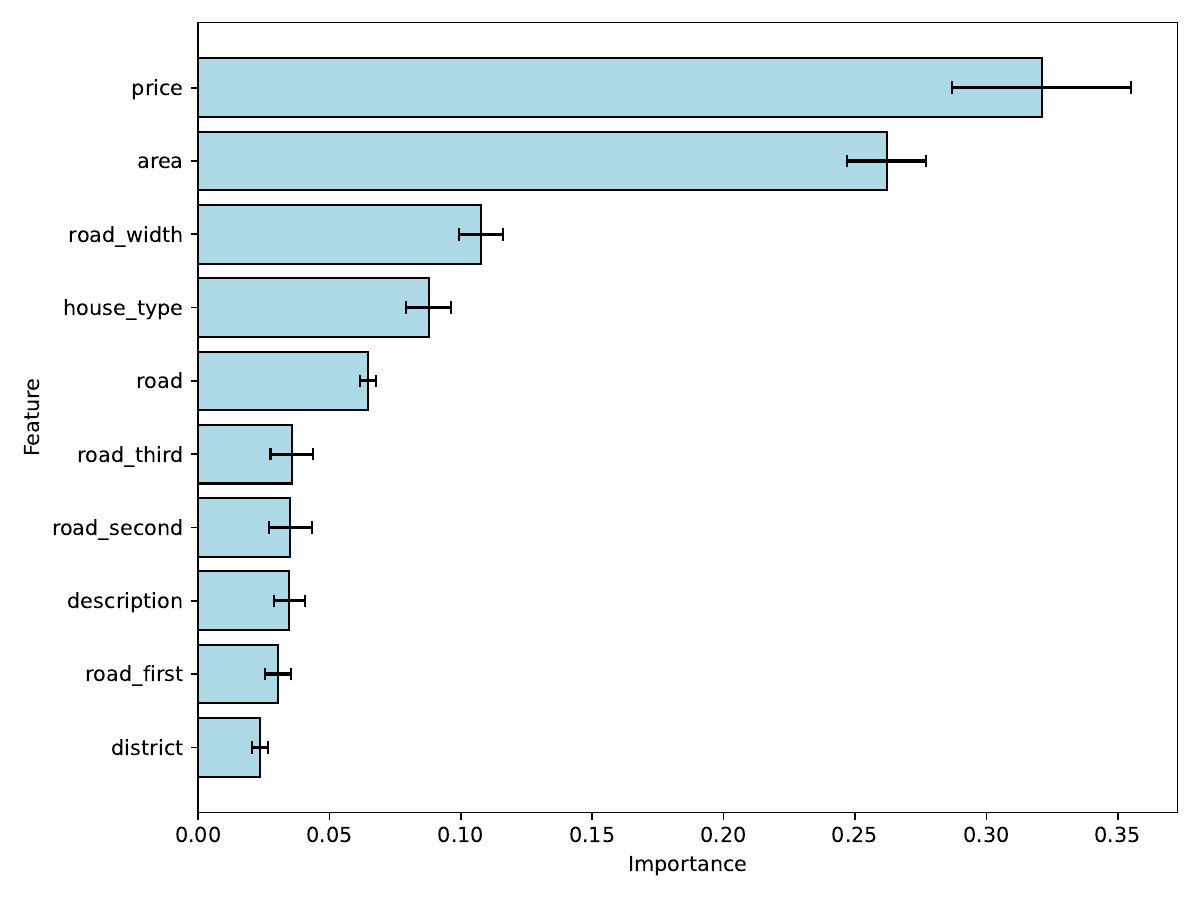}
\vspace{-0.3cm}
\caption{Feature importance scores and their standard deviations of our FADAML system on the test set.}
\label{fig:feature_importance}
\end{figure}

Using AutoGluon, we can also obtain the feature importance scores of the weighted ensemble model on our test set. These importance scores are computed using permutation-shuffling~\cite{breiman2001random}, where we randomly shuffle the values of each feature across all the examples and measure the drop in accuracy. This is a general technique for model interpretability that can be applied to various types of models. Figure~\ref{fig:feature_importance} plots the average importance scores of all features in FADAML together with their standard deviations from 5 random shuffle sets. From the figure, we see that the most important features are \emph{price} and \emph{area}, while the least important features are \emph{road\_first} and \emph{district} (see Table~\ref{tab:features} for the feature descriptions). Nevertheless, all features have a positive contribution to the performance of the system, as evidenced by their positive importance scores.

\subsection{Ablation Study}
\label{sec:ablation}

To help us understand the contributions of each component in FADAML to the whole system, we conduct an ablation study where we successively remove the following feature sets from the full system and measure its performance.

\begin{itemize}
  \item \textbf{Spatial multimodal features}: This feature set consists of the \emph{road\_first}, \emph{road\_second}, and \emph{road\_third} features obtained from the actual location of each property, as described in Section~\ref{label:feat-refine}.
  \item \textbf{Refined multimodal features}: This feature set consists of the other enriched and refined multimodal features in Section~\ref{label:feat-refine} (\emph{house\_type} and \emph{road\_width}).
  \item \textbf{Basic multimodal features}: This feature set consists of the basic multimodal features extracted from Section~\ref{sec:feat-extract} (\emph{price}, \emph{area}, \emph{road}, and \emph{district}).
\end{itemize}

After removing all the above multimodal features, we are left with the base model that uses only the \emph{description} text generated from Component 1 of our FADAML system (Section~\ref{sec:com1}). This description text will be automatically processed by AutoGluon to generate several generic natural language features discussed in Section~\ref{sec:automl} that will be used to train the base models.

\begin{table}[t]
\begin{center}
  \caption{Ablation study results for our FADAML system.}
  \vspace{-0.2cm}
  \label{tab:ablation}
  \small{
  \begin{tabular}{lcccc}
    \toprule
    Model & Precision & Recall & F1 & Accuracy \\
    \midrule
    Full model & 0.913 & 0.913 & 0.913 & 0.915 \\
    - Spatial multimodal features & 0.909 & 0.909 & 0.909 & 0.911 \\
    - Refined multimodal features & 0.878 & 0.878 & 0.878 & 0.881 \\
    - Basic multimodal features & 0.783 & 0.773 & 0.776 & 0.785 \\
    \bottomrule
  \end{tabular}
  }
\end{center}
\end{table}

In Table~\ref{tab:ablation}, we give the ablation study results for our system. The table shows that removing the spatial features slightly reduces the accuracy and F1 score of the system (from 91.5\% to 91.1\% for accuracy and from 91.3\% to 90.9\% for F1). When further removing the other refined multimodal features, the performance drops by around 3\% for all metrics. Finally, removing all the basic multimodal features degrades the performance of the system significantly, with at least 9.5\% reduction in all metrics. This result confirms the advantages of our FADAML system in extracting useful multimodal features for this problem. It is also worth noting that our base model without any multimodal features is still better than the baselines in Table~\ref{tab:compare_result}. This observation suggests that using AutoML is better for our problem than using pre-trained word embedding and deep learning models.

\subsection{Discussions}
\label{sec:discussion}

Despite the encouraging experiment results of our system, we note that there are still some limitations that need to be addressed to enhance its applicability in practice. First, the accuracy of $91.5\%$ (and consequently, the 9.8\% false positive rate and 7.4\% false negative rate), although better than the baselines, is still not sufficient for the deployment of the system to production. This is due to the difficulty of the problem relatively to the simplicity of our multimodal feature set. Thus, it is essential to extract more useful features to improve the performance of our system.

Another limitation of our study is the relatively moderate size of our dataset (with 29,085 examples) that could make it harder for deep learning approaches to work well. An increase in both size and geographical coverage of our dataset may help improve the accuracy of the neural network base models in our ensemble, thereby potentially improving the performance of the whole system. Additionally, noises in our dataset also pose a challenge for our system. Thus, it could also be useful to apply noise reduction techniques to improve the robustness of the system.

Finally, online fake advertisements can rapidly evolve, potentially outpacing the system's detection capabilities. Thus, we need to continuously retrain and update our system to keep up with the evolution of the problem. One potential solution to this problem is applying online learning~\cite{ben1997online, pham2015semi, nguyen2017online} and continual/life-long learning~\cite{luce, nguyen2024lifelong} to reduce the system maintenance cost.

\section{Conclusion}
\label{sec:conclusion}

We developed an end-to-end fake advertisement detector that combines multimodal and automated machine learning into a single system. Our method is capable of detecting fake advertisements on Vietnamese real estate websites with a high detection accuracy. The techniques in our system are very general and can be used in other applications. 

For future work, we aim to further improve our system by extracting more useful features from other data sources besides listing websites and adding more models into our AutoML component. For instance, we can incorporate information from images of the properties that can help us compare different properties visually. With the advances in computer vision, we can extract features from these images easily by using a pre-trained neural network and then add these features into our multimodal system.

Another potential future work is to extend our approach to other languages or other domains such as medical data, bio-informatics data, or other e-commerce domains. As our AutoML component is very general and flexible, it can be readily applied to these domains. However, a challenge when applying our system to other domains is the extraction of useful multimodal features for these domains. To solve this challenge, we may need to work with domain experts to understand the problems and determine which information would be important for each domain.

\subsection*{Acknowledgements}
We express our gratitude to Huynh Tan Phong and Ho Thi Yen Nhi for their support in acquiring and preprocessing data from public websites. We also extend our thanks to Vu Cao Can and his team for their contributions as real estate experts in our research.

\bibliographystyle{ieeetr}
\bibliography{main}

\end{document}